\documentclass[conference]{IEEEtran}
\IEEEoverridecommandlockouts
\usepackage{cite}
\usepackage{amsmath,amssymb,amsfonts}
\usepackage{algorithm}
\usepackage{algorithmic}
\usepackage{graphicx}
\usepackage{textcomp}
\usepackage{xcolor}
\usepackage{booktabs}
\usepackage{xurl}
\usepackage{caption}
\usepackage{tikz}
\usepackage{xurl}
\usetikzlibrary{arrows.meta,positioning,shapes.geometric}

\newcommand{\smallcaps}[1]{%
    {\fontsize{8}{9}\selectfont\textsc{#1}}%
}

\def\BibTeX{{\rm B\kern-.05em{\sc i\kern-.025em b}\kern-.08em
    T\kern-.1667em\lower.7ex\hbox{E}\kern-.125emX}}
\begin{document}

\title{Rethinking Decomposition: A Controlled Pilot Study of Agentic Text-to-SQL\\
}

\author{
\IEEEauthorblockN{
Duoduo Liao\textsuperscript{1},
Sehaj Gill\textsuperscript{1},
Shubham Sandeep Nirgude\textsuperscript{1},
Callie C. Liao\textsuperscript{2}
}
\IEEEauthorblockA{
\textsuperscript{1}School of Computing, George Mason University\\
{\{dliao2, sgill27, snirgude}\}@gmu.edu
}
\IEEEauthorblockA{
\textsuperscript{2}Department of Computer Science, Stanford University\\
ccliao@cs.stanford.edu
}
}

\maketitle

\begin{abstract}

Agentic Text-to-SQL systems often decompose complex questions into intermediate sub-questions before SQL generation. However, the independent contribution of decomposition remains unclear, as existing approaches typically combine it with other architectural components. To address this gap, we propose a controlled experimental framework to isolate the effects of decomposition and conduct a controlled pilot study using this framework. The evaluation compares 100 human-crafted natural-language question–gold SQL pairs from a real-world database and 62 additional questions from selected BIRD benchmark databases. Specifically, we compare direct generation with Large Language Model (LLM)-based manager-orchestrated decomposition under selective and forced routing, while keeping databases, schema, prompts, verification mechanisms, and evaluation protocol fixed across multiple model families. The results show no statistically significant performance improvement from manager-orchestrated decomposition under the evaluated conditions. Direct generation achieves the highest execution accuracy (90\% strict, 91\% normalized, 92\% LLM-judge-gated) while requiring substantially fewer inference calls. Although selective decomposition occasionally solves questions that direct generation fails, these improvements are offset by errors in other cases. Moreover, decomposition more than doubles end-to-end latency without improving accuracy. Human evaluation indicates that decomposition process generally produces valid intermediate sub-questions, suggesting that downstream SQL generation and composition remain key sources of error. These findings suggest that the decomposition should be applied selectively rather than assumed to provide consistent benefits, highlighting the need for more effective routing policies and SQL composition strategies to enhance the reliability and efficiency of agentic Text-to-SQL systems.

\end{abstract}


\section{Introduction}

Large Language Models (LLMs) have significantly advanced Text-to-SQL generation, enabling natural-language interfaces for databases \cite{11160657, 10.1145/3737873}. However, reliably translating complex user questions into executable SQL remains challenging, particularly for queries requiring multi-table reasoning, nested operations, and advanced SQL constructs. To address these challenges, recent agentic Text-to-SQL systems employ multi-stage workflows that incorporate decomposition, planning, refinement, and verification before generating final queries. These approaches build on the broader insight that decomposing complex tasks into structured intermediate steps can improve reasoning performance \cite{NEURIPS2023_72223cc6, wang-etal-2025-mac, ICLR2025_974ff7b5}.

However, the independent contribution of decomposition in Text-to-SQL remains unclear. Existing agentic systems typically combine decomposition with other components, including schema linking, retrieval, candidate generation, and self-correction. As a result, reported improvements cannot be directly attributed to decomposition alone. Moreover, introducing additional reasoning stages may lead to coordination and composition errors while increasing inference cost and latency \cite{NEURIPS2023_72223cc6, NEURIPS2025_b1041e52}. This raises a fundamental question: \textit{When does decomposition improve Text-to-SQL generation over direct query generation?}

To investigate this question, we propose a controlled experimental framework to isolate the effects of decomposition and conduct a pilot study within this framework. In this work, \emph{decomposition} refers to workflows that introduce intermediate sub-question generation and SQL composition stages between the user question and final SQL generation. Specifically, we compare direct generation with LLM-based manager-orchestrated decomposition under selective and forced routing, while keeping the database interface, schema representation, prompts, verification mechanisms, and evaluation protocol consistent across configurations. This study aims to investigate whether decomposition provides measurable end-to-end improvements and whether such improvements justify the associated computational overhead.

Our framework is evaluated on realistic relational databases with multiple interconnected tables, including a database derived from the U.S. Federal Highway Administration (FHWA) National Bridge Inventory (NBI) containing approximately 1.2 million rows across six tables \cite{fhwa2024portal} and two BIRD databases for cross-domain generalization \cite{NEURIPS2023_83fc8fab}. The evaluation includes 100 human-crafted natural-language question–gold SQL pairs for the NBI database and 62 additional questions from the BIRD databases. Multiple LLM families and model tiers are evaluated under three reasoning policies: direct generation, selective decomposition, and forced decomposition. The system measures execution accuracy, judge-based semantic equivalence, inference cost, and end-to-end latency.

Results show that manager-orchestrated decomposition does not consistently outperform direct generation under the evaluation conditions. Direct generation reaches above 90\% execution accuracies with OpenAI GPT-4o~\cite{openai2024gpt4o} while requiring substantially fewer inference calls. In contrast, forced decomposition increases latency by approximately 2--3$\times$ without reliable accuracy improvements. Human evaluation indicates that generated sub-questions are generally valid, suggesting that downstream SQL generation and composition remain important sources of error.

The main contributions are summarized as follows:
\begin{itemize}
    \item A controlled experimental framework and pilot study that isolates decomposition as an independent design choice in agentic Text-to-SQL, separating its effect from other architectural components.
    \item A systematic comparison of direct generation, selective decomposition, and forced decomposition across multiple LLM families and relational databases under a unified evaluation protocol, assessing both accuracy and computational cost.
    \item Analyses of routing behavior, sub-question validity through human evaluation, and failure modes, highlighting downstream SQL generation and composition as key remaining challenges.

\end{itemize}

\section{Related Work}

\emph{LLM-Based and Agentic Text-to-SQL.}
Large language models have advanced Text-to-SQL generation through improved prompting, schema grounding, retrieval, and staged generation. DIN-SQL decomposes Text-to-SQL into schema linking, difficulty classification, SQL synthesis, and self-correction~\cite{NEURIPS2023_72223cc6}. 
For in-context learning, DAIL-SQL investigates demonstration retrieval and ACT-SQL explores automated reasoning~\cite{10.14778/3641204.3641221, zhang-etal-2023-act}, whereas CodeS focuses on fine-tuning strategies for SQL generation~\cite{ 10.1145/3654930}. More recent agentic systems incorporate specialized components for decomposition, refinement, verification, and candidate selection~\cite{wang-etal-2025-mac, ICLR2025_974ff7b5}. However, these approaches typically combine decomposition with other architectural improvements, making its independent contribution to end-to-end performance difficult to establish.

\emph{Reasoning Through Decomposition.}
Decomposition has been widely explored as a strategy for improving LLM reasoning. Chain-of-Thought (CoT), least-to-most prompting, decomposed prompting, and plan-and-solve demonstrate the potential of structured intermediate reasoning for complex tasks~\cite{NEURIPS2022_9d560961, zhou2023least, wang-etal-2023-plan}. However, additional reasoning stages can increase computational cost and introduce opportunities for error propagation. Recent analyses of multi-agent LLM systems further show that failures may arise from coordination and integration between components rather than from individual reasoning steps alone~\cite{NEURIPS2025_b1041e52}. Most directly related to our work, \cite{tai-etal-2023-exploring} find that least-to-most styled question decomposition provides little benefit for Text-to-SQL parsing on Spider~\cite{yu-etal-2018-spider}, with finer-grained reasoning potentially increasing error propagation within a single-parser prompting framework. In contrast, our study investigates decomposition within an agentic pipeline comprising separate decomposer, solver, and composer components. We quantify its additional inference cost and use human evaluation of intermediate sub-questions to examine downstream failure sources. 

\emph{Evaluation of Agentic Systems.}
Evaluating LLM agents requires examining not only final accuracy but also intermediate behaviors, coordination efficiency, and computational cost. Benchmarks such as AgentBench ~\cite{ICLR2024_e9df36b2} and GAIA ~\cite{ICLR2024_25ae35b5} assess multi-step reasoning and tool-use capabilities, while recent studies highlight that coordination overhead and additional failure modes that can accompany multi-agent architectures~\cite{NEURIPS2025_b1041e52}. These findings motivate controlled evaluations that isolate individual architectural choices rather than comparing complete systems with multiple confounding factors. Accordingly, we investigate decomposition as a reasoning policy under a controlled framework, fixing the database, schema representation, prompts, models, verification procedure, and evaluation protocol. Our work complements agentic frameworks developed using the same FHWA data~\cite{gadiraju-etal-2025-conversational} and protocol-mediated multi-agent architectures~\cite{liao-etal-2025-agentmaster} by focusing on the independent contribution, computational overhead, and limitations of decomposition.

\section{Methodology}

\subsection{Overview}

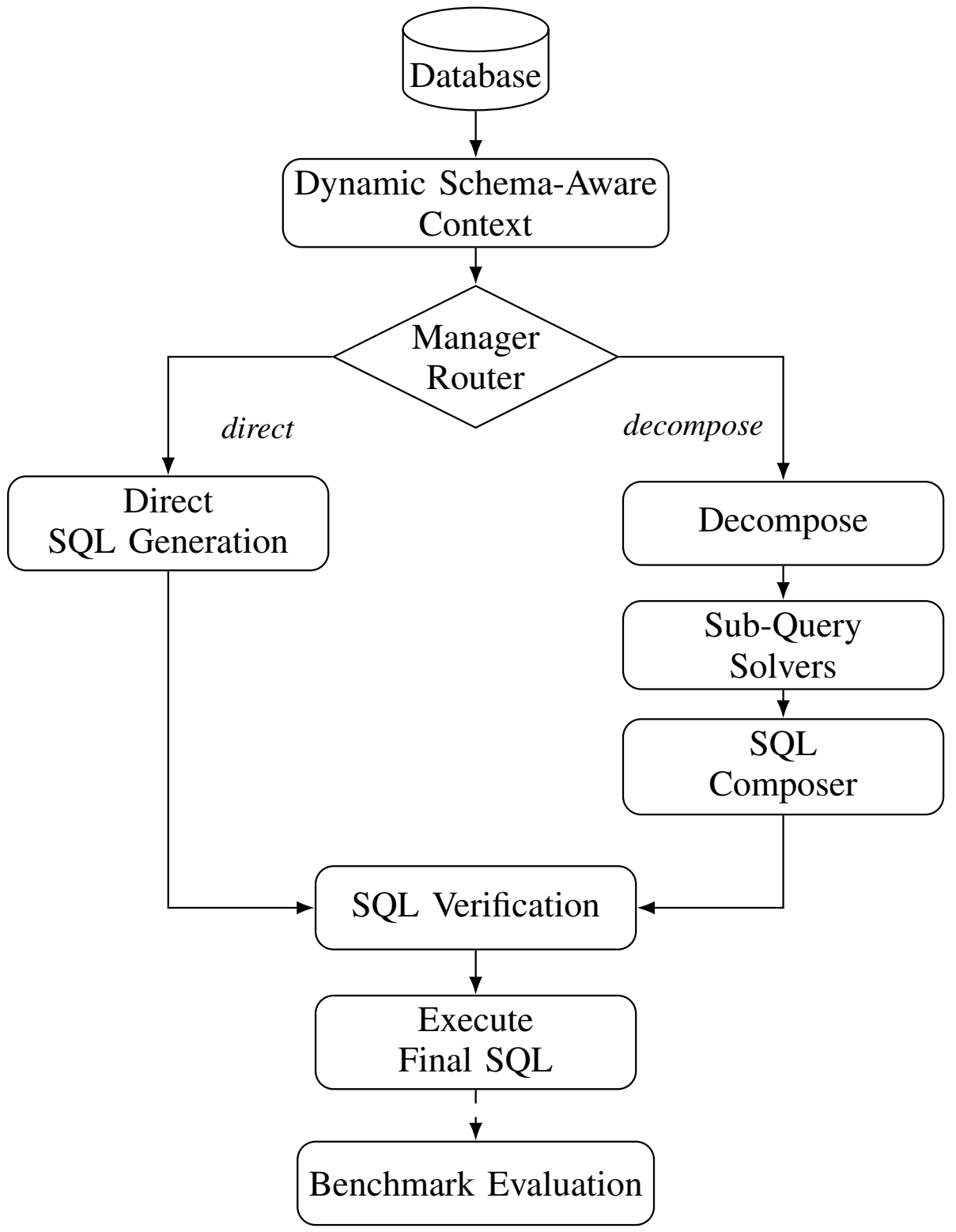
\begin{figure}[t]
\centering
\resizebox{0.9\columnwidth}{!}{%
\begin{tikzpicture}[
  font=\footnotesize, >={Latex[length=1.4mm]},
  box/.style={
    draw, rounded corners, align=center,
    minimum height=6.5mm, minimum width=25mm,
    inner sep=2.5pt
  },
  db/.style={
    draw, cylinder, shape border rotate=90,
    aspect=0.3, align=center,
    minimum height=8mm, inner sep=1.5pt
  },
  dia/.style={
    draw, diamond, aspect=2,
    align=center, inner sep=1pt
  },
]

% Top
\node[db] (db) at (0,0) {Database};

\node[box] (ctx) at (0,-1)
{Dynamic Schema-Aware\\Context};

\node[dia] (router) at (0,-2.2)
{Manager\\Router};

% Branches
\node[box] (direct) at (-2.4,-3.5)
{Direct\\SQL Generation};

\node[box] (dec) at (2.4,-3.5)
{Decompose};

\node[box] (solve) at (2.4,-4.45)
{Sub-Query\\Solvers};

\node[box] (compose) at (2.4,-5.4)
{SQL\\Composer};

% Shared verification and execution
\node[box] (verify) at (0,-6.5)
{SQL Verification};

\node[box] (exec) at (0,-7.55)
{Execute\\Final SQL};

\node[box, fill=black!0] (eval) at (0,-8.65)
{Benchmark Evaluation};

% Edges
\draw[->] (db)--(ctx);
\draw[->] (ctx)--(router);

% Routing
\draw[->] (router)-|(direct);
\draw[->] (router)-|(dec);

% Decomposition workflow
\draw[->] (dec)--(solve);
\draw[->] (solve)--(compose);

% Merge into verification
\draw[->] (direct)|-(verify);
\draw[->] (compose) |-(verify);

% Verification pipeline
\draw[->] (verify)--(exec);
\draw[->, dashed] (exec)--(eval);

% Labels
\node[font=\scriptsize] at (-1.7,-2.75) {\textit{direct}};
\node[font=\scriptsize] at (1.7,-2.75) {\textit{decompose}};

\end{tikzpicture}
}
\caption{Overview of the controlled framework. All reasoning policies share identical database context and schema construction. The LLM-based manager routes queries to direct generation or an orchestrated decomposition workflow involving sub-question generation, solving, and SQL composition. All final queries are evaluated under the same verification and benchmark metrics (strict, normalized, and LLM-judge execution accuracies).}
\label{fig:arch}
\end{figure}

This pilot study introduces a controlled experimental framework for investigating decomposition as a reasoning policy in agentic Text-to-SQL systems (Figure~\ref{fig:arch}). Given a database and a natural-language question, the framework constructs question-specific schema context, generates SQL through direct generation or LLM-based manager-orchestrated decomposition, applies a shared verification procedure, and executes the resulting query.

The reasoning policy is the primary experimental variable. Direct generation serves as the baseline, while decomposition introduces additional stages for sub-question generation, solving, and SQL composition. To isolate the contribution of decomposition, all policies use the same database interface, schema representation, prompts, LLMs, verification procedure, and evaluation protocol.

%-------------------------
%  DeComSQL Flowchart
%-------------------------------------

%-------------------------

\subsection{Schema Construction}
\label{sec:schema}

This study employs dynamic schema-context construction to provide question-specific database information for SQL generation. For each natural-language question, the schema module retrieves relevant tables, columns, and relationships to construct a schema context shared across all reasoning policies. This design ensures that both direct generation and decomposition-based approaches receive identical schema information, allowing performance difference to be attributed to the reasoning policy rather than variations in schema grounding.

\subsection{Reasoning Policies}

Direct generation is compared with manager-orchestrated decomposition under the same schema context, verification procedure, and execution pipeline. In this study, \emph{decomposition} refers to the introduction of intermediate sub-question generation, sub-query solving, and SQL composition stages between the original question and the final SQL generation.

\subsubsection{Direct Generation}

Direct generation serves as the baseline policy. A single agent maps the original question and its question-specific schema context directly to one SQL query without explicit intermediate decomposition. Schema interpretation, reasoning, and query synthesis are performed within a single generation step. The resulting query then passes through the shared verification and execution pipeline.

\subsubsection{Manager-Orchestrated Decomposition}

The decomposition policy introduces an LLM-based manager that routes questions through a multi-stage workflow consisting of decomposition, solving, and composition. The decomposer generates intermediate sub-questions and their dependency relationships. Solver agents generate SQL queries for individual sub-questions, and a composer integrates these intermediate solutions into one final SQL query. The generated query then passes through the same verification and repair pipeline as direct generation. For composition retries, an optional stronger model is used. The initial composition attempt always uses the same model as the decomposer and solver agents. The stronger model is invoked only when composed query is rejected and requires recomposition. 

Three routing conditions are evaluated: \emph{direct}, where all questions are solved without decomposition; \emph{selective decomposition}, where the manager decomposes questions requiring two or more distinct computational stages; and \emph{forced decomposition}, where the manager is instructed to decompose by default, reserving single-step plans for trivial lookups. The selective and forced policies use identical prompts except for the routing policy instructions.
%(Figure~\ref{fig:prompt-decompose} in Appendix~\ref{app:prompts}).
Although forced routing increases decomposition frequency, the manager still makes the final routing decision and may produce single-step plans. In practice, the two policies differ substantially as forced routing decomposes 72--97 of the 100 questions, compared with 10--36 under selective routing (Table~\ref{tab:headline}).

\subsection{Verification and Execution}

Generated SQL queries from all policies pass through the same SQL validation and repair procedure before execution. The validator checks SQL validity, executes the query, and provides execution feedback for up to two repair attempts. Decomposition-based policies additionally apply reference-free semantic checks to intermediate sub-questions and final query alignment. These checks do not access gold SQL or reference answers.

After validation and any necessary regeneration attempts, the final SQL query is executed against the database and evaluated using the shared benchmark protocol. Applying the same verification, repair, execution, and evaluation procedures across policies ensures that observed performance differences primarily reflect the reasoning strategy rather than variations in downstream processing.

%--------------------------
% ==== Table 1: Comprehensive results (wide, spans both columns) ====
\begin{table*}[t]
\centering
\small
\setlength{\tabcolsep}{5pt}
\begin{tabular}{lllccccccc}
\toprule
 & & & & & \multicolumn{5}{c}{Accuracy (\%)} \\
\cmidrule(lr){6-10}
Policy & Base & Retry & Route & $n_{\text{dec}}$ 
& EX$_{\text{strict}}$ & EX$_{\text{norm}}$ & +Gated & +All Judge & Avg \\
\midrule
D & gpt-4o        & --           & --  & --  & \textbf{90} & \textbf{91} & \textbf{92} & \textbf{93} & \textbf{91.5} \\
D & gpt-4.1       & --           & --  & --  & 83 & 89 & 89 & 89 & 86.0 \\
D & gpt-4.1-mini  & --           & --  & --  & 86 & 88 & 88 & 88 & 87.0 \\
D & gpt-4o-mini   & --           & --  & --  & 82 & 82 & 84 & 85 & 83.5 \\
\addlinespace
M & gpt-4o        & --           & SD & 36  & 85 & 87 & 88 & 89 & 87.0 \\
M & gpt-4o        & --           & FD & 85  & 86 & 87 & 88 & 89 & 87.5 \\
M & gpt-4.1       & gpt-4o       & SD & 23  & 80 & 89 & 89 & 90 & 85.0 \\
M & gpt-4.1       & gpt-4o       & FD & 84  & 76 & 87 & 88 & 89 & 82.5 \\
M & gpt-4.1-mini  & gpt-4o       & SD & 10  & 85 & 88 & 88 & 88 & 86.5 \\
M & gpt-4.1-mini  & gpt-4o       & FD & 72  & 85 & 88 & 89 & 90 & 87.5 \\
M & gpt-4o-mini   & gpt-4.1-mini & SD & 79  & 79 & 80 & 82 & 85 & 82.0 \\
M & gpt-4o-mini   & gpt-4.1-mini & FD & 97  & 73 & 77 & 79 & 80 & 76.5 \\
\bottomrule
\end{tabular}
\caption{
Accuracy on the 100-question NBI benchmark under dynamic schema construction.
\textbf{Policy}: D = direct generation, M = LLM-based manager-orchestrated decomposition.
\textbf{Route}: SD = selective decomposition, FD = forced decomposition.
$n_{\text{dec}}$ reports the number of questions routed through decomposition.
EX$_{\text{strict}}$ is strict execution accuracy (1\% numeric tolerance, columns matched by name); EX$_{\text{norm}}$ applies the same comparison with value-based column matching and a 5\% tolerance to account for column aliasing and reordering. Both are deterministic comparators.
+Gated only adds judge rescues that the row judge confirms or the semantic judge attributes to a mechanical mismatch, while +All Judge adds every accepted rescue and so upper-bounds the judge contribution.
\textbf{Avg} is the mean of EX$_{\text{strict}}$ and +All Judge.
}
\label{tab:headline}
\end{table*}
%--------------------------

\subsection{Evaluation Protocol}
\label{sec:eval}

The framework is evaluated using accuracy and efficiency metrics under a consistent scoring procedure across all reasoning policies. Gold SQL or reference answers are withheld from the generation and verification stages and only reserved for evaluation. The evaluation consists of execution-based correctness, judge-gated semantic recovery, and computational efficiency.

\subsubsection{Execution Accuracy}
Execution accuracy is the primary evaluation metric, measuring the percentage of questions for which the generated SQL produces the same result as the gold SQL when executed on the database. Result comparison is permutation-invariant, which ignores row ordering, tolerates column aliasing and column ordering differences, and allows an $\alpha$\% relative tolerance for numerical values. This pilot study reports two execution accuracy: i) \emph{Strict execution accuracy}, EX$_{\text{strict}}$, uses column-name matching with 1\% relative tolerance for numerical values; and ii) \emph{Normalized execution accuracy}, EX$_{\text{norm}}$, applies the same comparison with value-based column matching with a 5\% relative tolerance, accommodating difference in column aliases and ordering. Both metrics use deterministic result comparators. Questions with non-executable gold SQL are excluded from evaluation across all policies because no valid reference result can be obtained. 

\subsubsection{Judge-Gated Accuracy}
Judge-gated accuracy extends execution accuracy by recovering cases that fail strict comparison but may be semantically equivalent to the gold answer. It applies two LLM-based rescue checks \cite{NEURIPS2023_91f18a12}: i) a result level judge, which compares the generated and gold result sets in the context of the natural-language question; and ii) an SQL judge, which assesses the semantic equivalence of the generated and gold queries. A failed case is recovered only when the result-level judge confirms that the generated results correctly answer the question, or when the SQL judge confirms query equivalence and the remaining differences fall into predefined categories, such as column ordering, additional projected columns, alias naming, value formatting or row ordering. Judge-based recovery applies only to cases that fail strict execution comparison and never overrides an execution-correct prediction. Both measures are reported separately. To reduce self-agreement bias, the judge model is selected from a different model family than the SQL generator. 

\subsubsection{Efficiency Measurement}
Computational efficiency is evaluated using LLM call volume and end-to-end latency to quantify the computational overhead introduced by additional reasoning stages. Since verification, repair, execution, scoring, and cost accounting are consistent across all conditions, these metrics capture the computational trade-offs associated with each reasoning policy.

%=========================================
\section{Experimental Results and Analysis}
\label{sec:results}

This section presents a pilot evaluation of decomposition under controlled reasoning policies. It compares direct generation with selective and forced decomposition, examining routing behavior, failure modes, computational cost, and preliminary cross-domain generalization.

%%%%%%%%%%%%%%%%%%%%%%%%%
\begin{figure*}[t]
\centering
\includegraphics[width=0.8\textwidth]{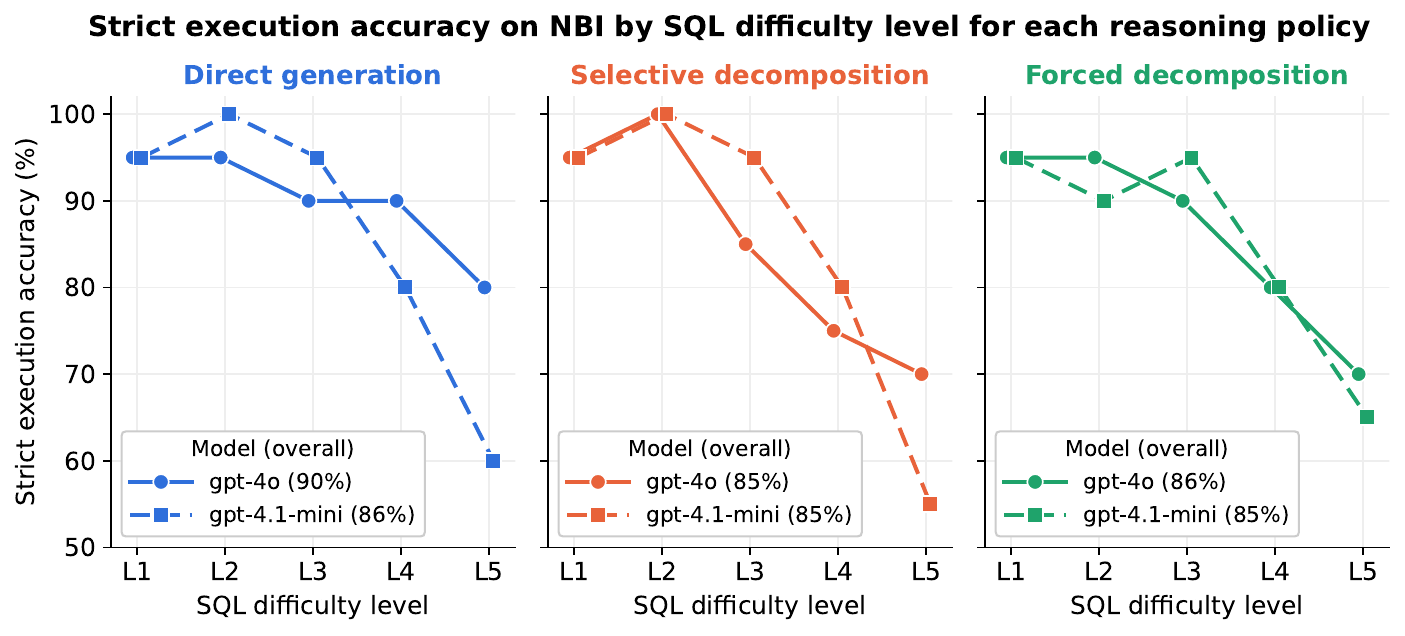}
\caption{Strict execution accuracy, EX$_{\text{strict}}$, (\%) on NBI by SQL difficulty level (20 questions per level; L1 is the simplest and L5 the most complex), with one panel per reasoning policy. %Forced decomposition is tested only with GPT-4.1-mini to assess its benefits for a smaller model on complex queries.
}
\label{fig:bylevel}
\end{figure*}
%%%%%%%%%%%%%%%%%%%%%%%%%

%---------------------------------------
\subsection{Experimental Setup}
\label{sec:setup}

% addressed reviewer comment on why NBI data
\subsubsection{Benchmarks}
The primary evaluation uses a benchmark derived from the FHWA's NBI. 
NBI provides a realistic testbed for analytical Text-to-SQL workloads, combining a large, normalized government database with domain-specific classification codes maintained in lookup tables, reflecting characteristics of enterprise databases. Furthermore, its specialized domain is relatively underrepresented in existing public Text-to-SQL benchmarks. The benchmark contains 100 hand-crafted natural-language questions with corresponding gold SQL queries over a six-table relational database, including \smallcaps{bridges} (569,864 records) and \smallcaps{inspections} (624,193 records). Questions were manually authored and verified by database experts, and the gold SQL queries and their execution results were checked for consistency. The questions are evenly distributed across five SQL difficulty levels, from single-table aggregation (L1) to window functions and common table expressions (L5). Appendix~\ref{app:Benchmark} provides detailed database statistics. Additional experiments on two BIRD databases, \smallcaps{california\_schools} (30 questions) and \smallcaps{financial} (32 questions), provide cross-domain evaluation. 

\subsubsection{Models and Policies}
The NBI experiments evaluate four OpenAI models: GPT-4o, GPT-4.1, GPT-4.1 mini, and GPT-4o-mini~\cite{openai2024gpt4o,openai2025gpt41}, with Claude Sonnet 4.6 serving as the judge for judge-gated accuracy. For BIRD, the model-family roles are reversed: Claude models~\cite{anthropic2024claude3,anthropic2025claude4} generate SQL and OpenAI models serve as judges. All models are evaluated under three reasoning policies: direct generation, selective decomposition, and forced decomposition. Direct generation produces SQL directly from the question and schema, while manager-orchestrated policies introduce decomposition, sub-query solving, and composition stages.

\subsubsection{Evaluation Protocol}
All policies use consistent databases, schema representations, prompts, and verification procedures. Performance is measured using execution accuracy and judge-gated accuracy, while efficiency is measured through LLM call volume and end-to-end latency. Statistical significance is evaluated using McNemar's exact test on paired strict execution accuracy with 95\% confidence intervals reported for paired accuracy differences.

\subsubsection{Human Evaluation}
Human evaluation examines the quality of sub-questions generated by forced decomposition. It covers 258 sub-questions from 82 decomposed queries, assessing their validity and relevance to the original questions. Detailed evaluation procedures and inter-annotator agreement statistics are provided in Appendix~\ref{app:human}.

%%%%%%%%%%%%%%%%%%%%%%%%%%%%%%%%%%%%%%%%%%%
\begin{figure*}[t]
\centering
\includegraphics[width=1\textwidth]{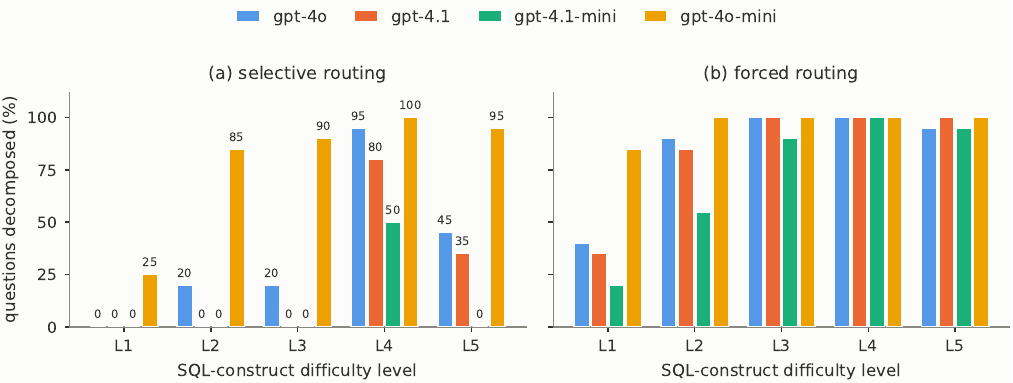}
\caption{Decomposition routing by SQL difficulty level across models (20 questions per level). A question is counted as decomposed when its generated plan contains multiple sub-questions. \textbf{(a)} Selective routing. \textbf{(b)} Forced routing (decompose by default). Since routing remains model-driven, forced routing does not achieve full coverage, decomposing 72--97 of 100 questions depending on the model.}

\label{fig:routing_by_level_dynamic}
\end{figure*}
%%%%%%%%%%%%%%%%%%%%%%%%%%%%%%%%%%%%%%%%%%%
%----------------------------------------
\subsection{Overall Comparison of Reasoning Policies}
\label{sec:overall}

Table~\ref{tab:headline} compares direct generation with manager-orchestrated decomposition under selective and forced routing. Direct generation achieves the strongest overall performance, with GPT-4o reaching 90\% strict execution accuracy and 93\% judge-gated accuracy.

\subsubsection{Limited Gains from Decomposition}
Neither selective nor forced decomposition yields a statistically significant improvement over direct generation. Selective decomposition underperforms the direct baseline across all models. For GPT-4o, for example, strict accuracy decreases from 90\% with direct generation to 85\% with selective decomposition. Forced decomposition increases decomposition coverage but does not improve accuracy, with the best configurations reaching 86\% strict accuracy. These findings suggest that additional reasoning stages do not consistently translate into better SQL generation under the evaluated conditions.

\subsubsection{Effects of Judge-Gated Evaluation}
Judge-based evaluation recovers semantically equivalent outputs that fail strict comparison, increasing accuracy by 2-4 percentage points across configurations. However, these recoveries do not change the relative ranking of reasoning policies, with direct generation remaining the strongest approach.

\subsubsection{Model Capability Versus Additional Reasoning}
Performance improvements are more consistently associated with stronger base models than additional decomposition stages. Direct generation improves from 82\% accuracy with GPT-4o-mini to 90\% with GPT-4o, whereas selective decomposition improves from 79\% to 85\%. These results suggest that, under the evaluated configurations, increasing model capability provides more consistent performance gains than introducing additional coordination stages.

%-----------------------------------------
\subsection{Routing Behavior and Effectiveness}
\label{sec:routing}

\subsubsection{Complexity-Aware Routing}
Figure~\ref{fig:routing_by_level_dynamic}a shows that selective routing generally aligns with SQL difficulty. Stronger models concentrate decomposition on L4--L5 queries and rarely decompose simpler ones, whereas GPT-4o-mini applies decomposition more broadly.  Decomposition is primarily assigned to structurally complex queries involving nested sub-queries, window functions, and common table expressions. However, overall coverage remains limited, with only 10--36 questions decomposed depending on the model (Table~\ref{tab:headline}).

\subsubsection{Routing Effectiveness}

Despite aligning decomposition strategies with query complexity, difficulty-aware routing does not consistently improve execution accuracy. Figure~\ref{fig:bylevel} shows relatively comparable performance across policies on L1--L3 (from 85\% to 100\%) but large differences emerge on more difficult queries (L4--L5). 
For GPT-4.1 mini, forced decomposition improves L5 accuracy from 60\% to 65\%, while selective decomposition reduces to 55\%. However, overall accuracy remains slightly lower than direction generation (85\% vs 86\%). For GPT-4o, both decomposition strategies reduce L5 accuracy from 80\% to 70\% and overall accuracy from 90\% to 85\% and 86\%, respectively. These results suggest that decomposition provides limited benefits on complex queries and does not consistently improve overall performance.

\subsubsection{Limits of Complexity-Based Routing}
Increasing decomposition coverage does not improve performance. Forced routing decomposes 72--97 questions depending on the model (Figure~\ref{fig:routing_by_level_dynamic}b; Table~\ref{tab:headline}), yet does not outperform direct generation. These results show that query complexity is insufficient to predict when decomposition will improve SQL generation. Effective routing must therefore consider the expected benefit of additional reasoning rather than relying only on complexity.

%----------------------------------
\subsection{Reasoning Policy Complementarity}
\label{sec:agreement}

Table~\ref{tab:confusion} compares direct generation and manager-orchestrated decomposition at the question level using strict execution accuracy, highlighting recovered questions and newly introduced errors.

\subsubsection{Selective Decomposition}
Selective decomposition shows strong agreement with direct generation. Both policies produce correct SQL for 84 questions and fail on the same 13 questions, agreeing on 97 of 100 questions. Decomposition recovers one additional question but introduces two new failures, providing limited complementary benefit.

\subsubsection{Forced Decomposition}
Forced decomposition increases disagreement without improving overall accuracy. It corrects five questions incorrectly answered by direct generation, but introduces six failures on questions where direct generation succeeds, leaving the manager correct on 85 questions despite substantially increasing decomposition usage.

\subsubsection{Paired Question-Level Analysis}
Since all policies are evaluated on the same questions, comparisons are paired at the question level. Most predictions are identical across policies (97\% for selective routing and 89\% for forced routing), with aggregate performance differences driven by the relatively few questions whose outcomes change. Overall, decomposition occasionally recovers failures from direct generation, but these gains are not consistently larger than the failures it introduces.

\subsubsection{Statistical Significance}
McNemar's exact tests indicate no significant difference between direct generation and either policy. For GPT-4.1 mini, selective and forced decomposition both reduce accuracy by 1 percentage point with (95\% confidence interval of $[-4.4, +2.4]$) and ($[-7.5, +5.5]$) relative to direct generation, respectively. These intervals show that decomposition does not produce a large improvement in accuracy.

Overall, decomposition provides limited complementary benefits, with recovered failures offset by newly introduced errors. 

%---------------------------------------

%---------------------------------------
% ==== Table 3: Direct vs. manager per-question effect ====
\begin{table}[t]
\centering
\small
\begin{tabular}{lccc}
\toprule
\multicolumn{4}{c}{Selective routing: 10 decomposed, all at level~4} \\
\midrule
Questions ($n{=}100$) & Mgr correct & Mgr wrong & Total \\
Direct correct & 84 & 2  & 86 \\
Direct wrong   & 1  & 13 & 14 \\
\cmidrule(l){1-4}
Total          & 85 & 15 & 100 \\
\midrule
\multicolumn{4}{c}{Forced routing: 72 decomposed, GPT-4o composer retry} \\
\midrule
Questions ($n{=}100$) & Mgr correct & Mgr wrong & Total \\
Direct correct & 80 & 6 & 86 \\
Direct wrong   & 5  & 9 & 14 \\
\cmidrule(l){1-4}
Total          & 85 & 15 & 100 \\
\bottomrule
\end{tabular}
\caption{
Per-question agreement between direct generation and manager policies on 100 questions with GPT-4.1-mini under strict execution accuracy for selective and forced decomposition. For forced routing, GPT-4o, a more capable model, is used to recompose the output if the initial attempt fails.
%Rows/columns denote outcomes; diagonal cells indicate agreement. 
}
\label{tab:confusion}
\end{table}
%---------------------------------------

%---------------------------------------
\subsection{Failure Analysis}
\label{sec:failure}

This section examines whether decomposition failures originate from intermediate reasoning or downstream SQL generation and composition.

\subsubsection{Intermediate Decomposition Quality}
Human evaluation indicates that invalid decomposition is not the primary source of errors. Across 258 generated sub-questions from 82 forced-decomposition questions, 96.9\% were judged valid, and only 92.7\% of decomposed questions contained valid sub-questions (Table~\ref{tab:decomp-quality}). This indicates that the manager generally produces meaningful intermediate tasks, although valid decomposition does not guarantee correct final SQL. Errors may still arise during intermediate query generation, integration, or scoping.

\subsubsection{Surface-Form Divergence}
Some failures arise from output differences rather than semantic errors. The composition stage may introduce additional columns or alternative projections that fail the strict execution comparison. For example, a query asking ``which states have an average bridge age older than the national average'' returns an additional \smallcaps{state\_code} column. Although semantically equivalent, this result fails strict comparison and is recovered through judge-gated evaluation.

\subsubsection{Semantic Errors in Composition}
Other failures involve incorrect global semantics that persist after judge evaluation. For example, a question requesting the top 15 states by bridge count together with rankings by average bridge age and total deck area requires rankings to be computed across all states before selecting. Instead, the manager computes rankings after filtering to the top 15 states, producing an incorrect result. Although the decomposition identifies the necessary sub-tasks, the composition stage fails to preserve global semantics.

Overall, these findings suggest that decomposition quality is not the primary bottleneck. The greater challenge lies in generating and composing intermediate SQL solutions while preserving global semantics.

%--------------------------------------
\subsection{Accuracy-Efficiency Trade-off}
\label{sec:cost}

Table~\ref{tab:cost} compares inference cost across reasoning policies using GPT-4.1 mini. Direct generation requires one generation call per question with an average latency of 6.1 seconds. Selective and forced decomposition increase latency to 12.8 seconds (2.1$\times$) and 20.3 seconds (3.3$\times$), respectively, due to additional routing, sub-question generation, SQL generation, composition, and verification stages. However, the computational overhead does not yield consistent accuracy improvements, indicating an unfavorable accuracy-cost trade-off under the evaluated settings. These results suggest applying decomposition selectively rather than as a default Text-to-SQL pipeline.

%-------------------------

% ==== Table 4: Cost ====
\begin{table}[!th]
\centering
\small
\setlength{\tabcolsep}{4pt}
\begin{tabular}{lccl}
\toprule
Configuration & Latency & vs.\ direct & Gen. calls/ques. \\
\midrule
Direct           & 6.1\,s  & 1.0$\times$ & 1 \\
Manager (Selective)   & 12.8\,s & 2.1$\times$ & $\geq$2 (+ decomp.) \\
Manager (Forced) & 20.3\,s & 3.3$\times$ & $\sim$$N{+}2$ \\
\bottomrule
\end{tabular}
\caption{
Inference cost comparison on the NBI using GPT-4.1-mini. 
%Decomposition adds 2-3$\times$ latency without consistent accuracy gains. Selective routing reduces overhead, while forced routing incurs the highest cost.
}
\label{tab:cost}
\end{table}
%-------------------------

%----------------------------
%----------------------------
\subsection{Cross-Domain Generalization}
\label{sec:generalization}

To assess generalization, all three reasoning policies are evaluated on two BIRD databases, reversing the model-family roles by using Claude generators and an OpenAI judge. Table~\ref{tab:bird} shows that NBI findings extend across additional database domains and model families. Direct generation with Claude Sonnet achieves the highest accuracy on both databases, while decomposition matches or underperforms its direct-generation baseline across all configurations. Model capability dominates routing policy: replacing Claude Haiku with Claude Sonnet under direct generation improves strict execution accuracy by 13.3 and 15.6 percentage points on the two databases, respectively, exceeding the gains associated with any routing change. 

A stronger composer also improves decomposition performance: on \smallcaps{california\_schools}, using a stronger model for composition raises Claude Haiku's accuracy from 36.7\% to 43.3\%, although it remains below direct generation with Claude Sonnet. This suggests that additional model capability at the composition stage can partially mitigate decomposition errors, but does not eliminate the overall advantage of direct generation.

% BIRD results added from appendix

% ==== Table 6: BIRD robustness (Claude generators, OpenAI judge) ====
\begin{table*}[t]
\centering
\small
\setlength{\tabcolsep}{5pt}
\begin{tabular}{lllccccccc}
\toprule
 & & & & & \multicolumn{5}{c}{Execution accuracy (\%)} \\
\cmidrule(lr){6-10}
Policy & Base & Retry & Route & $n_{\text{dec}}$ & EX$_{\text{strict}}$ & EX$_{\text{judge}}$ & +Gated & +All Judge & Avg \\
\midrule
\multicolumn{10}{l}{\textsc{california\_schools} (30 questions)} \\
D & sonnet & --     & --  & --  & \textbf{50.0} & \textbf{56.7} & \textbf{76.7} & \textbf{80.0} & \textbf{65.0} \\
D & haiku  & --     & --  & --  & 36.7 & 50.0 & 56.7 & 66.7 & 51.7 \\
M & sonnet & --     & SD & 6   & 43.3 & 50.0 & 73.3 & 73.3 & 58.3 \\
M & sonnet & --     & FD & 29  & 43.3 & 50.0 & 63.3 & 73.3 & 58.3 \\
M & haiku  & --     & SD & 18  & 36.7 & 50.0 & 56.7 & 66.7 & 51.7 \\
M & haiku  & --     & FD & 29  & 33.3 & 36.7 & 50.0 & 66.7 & 50.0 \\
M & haiku  & sonnet & SD & 18  & 43.3 & 50.0 & 60.0 & 63.3 & 53.3 \\
M & haiku  & sonnet & FD & 29  & 36.7 & 40.0 & 53.3 & 66.7 & 51.7 \\
\addlinespace
\multicolumn{10}{l}{\textsc{financial} (32 questions)} \\
D & sonnet & --     & --  & --  & \textbf{62.5} & \textbf{68.8} & \textbf{68.8} & \textbf{78.1} & \textbf{70.3} \\
D & haiku  & --     & --  & --  & 46.9 & 53.1 & 59.4 & 62.5 & 54.7 \\
M & sonnet & --     & SD & 14  & 50.0 & 59.4 & 62.5 & 62.5 & 56.3 \\
M & sonnet & --     & FD & 32  & 53.1 & 59.4 & 59.4 & 65.6 & 59.4 \\
M & haiku  & --     & SD & 20  & 37.5 & 43.8 & 46.9 & 56.2 & 46.9 \\
M & haiku  & --     & FD & 32  & 37.5 & 46.9 & 46.9 & 56.2 & 46.9 \\
M & haiku  & sonnet & SD & 19  & 37.5 & 43.8 & 50.0 & 59.4 & 48.5 \\
M & haiku  & sonnet & FD & 32  & 37.5 & 40.6 & 40.6 & 56.2 & 46.9 \\
\bottomrule
\end{tabular}
\caption{
Cross-Domain Evaluation on BIRD. Generalization evaluation on two databases from the BIRD benchmark using Claude-based generators and an OpenAI judge. 
\textbf{Policy}: D = direct generation, M = manager-orchestrated decomposition.
\textbf{Retry} indicates the stronger model used for composer retry when enabled; ``--'' indicates no retry.
\textbf{Route}: SD = selective decomposition, FD = forced decomposition.
$n_{\text{dec}}$ reports the number of questions routed through decomposition.
EX$_{\text{strict}}$ denotes strict execution accuracy; EX$_{\text{judge}}$, +Gated, and +All Judge report accuracy under increasingly permissive judge-based equivalence criteria.
}
\label{tab:bird}
\end{table*}

%=============================
\section{Discussion}
\label{sec:discussion}

\emph{Decomposition and End-to-End Performance.}
Decomposition often produces valid sub-questions, but this does not guarantee correct intermediate SQL or final query composition. Additional reasoning stages introduce opportunities for errors that can offset the benefits. These findings highlight downstream SQL generation and composition as important challenges in agentic pipelines.

\emph{Conditional Benefits of Decomposition.}
The effectiveness of decomposition depends on model capability and query complexity. It may benefit weaker models on complex queries, while stronger models perform more consistently through direct generation. Cross-domain experiments also suggest that stronger composition capabilities can mitigate decomposition errors. However, decomposition offers no consistent advantage, as its effectiveness depends on the ability of models to solve and integrate intermediate tasks reliably.

\emph{Implications for Agentic Text-to-SQL Design.}
Decomposition should not be treated as a default component of agentic Text-to-SQL systems. Direct generation offers a more efficient approach when models can adequately handle queries, whereas decomposition may be useful when its expected benefits justify additional inference costs. Effective routing should consider not only query complexity but also the likelihood that additional reasoning improves the final result. Improving SQL composition and preserving global query semantics remain important directions for more reliable agentic systems.

%=============================

%=============================
\section{Conclusion}
\label{sec:conclusion}

This work proposes a controlled experimental framework for evaluating manager-orchestrated decomposition as a reasoning policy in agentic Text-to-SQL and conducts a pilot study using this framework. This framework isolates the decomposition effects by holding the database interface, schema representation, verification procedure, and evaluation protocol fixed.
Under the evaluated settings, decomposition does not significantly outperform direct generation, which achieves the highest execution accuracy (90\% strict, 91\% normalized, and 92\% LLM-judge-gated), compared with 85\% for selective and 86\% for forced decomposition under strict execution accuracy. Although decomposition occasionally corrects errors made by direct generation, these gains are offset by newly introduced failures. Moreover, decomposition increases inference cost, resulting in a $2$--$3\times$ increase in end-to-end latency without consistent accuracy improvements. Human evaluation indicates that intermediate sub-questions are generally valid, suggesting that downstream SQL generation and composition remain key sources of error. 
Additional experiments on BIRD provide preliminary support for these findings across database domains and model families. Overall, decomposition should be applied only when its expected benefits justify the computational cost and potential failure risks, motivating more effective routing and SQL composition strategies. 

%=============================
\section*{Limitations}

This pilot study provides a controlled analysis of reasoning policies for agentic Text-to-SQL, but several limitations define the scope of the conclusions.

\emph{Benchmark Scope.}
The primary evaluation uses 100 natural-language questions over a relational database derived from the FHWA's NBI. Although the benchmark covers SQL patterns from simple aggregation to window functions and common table expressions, it does not represent the full diversity of enterprise-scale analytical workloads. Experiments on two BIRD databases provide additional cross-domain evidence, but broader evaluations are needed to establish generalizability.

\emph{Schema and Architecture Scope.}
The study maintains a consistent schema-context construction across reasoning policies to isolate the decomposition effect. However, larger databases may introduce additional challenges in schema retrieval, context selection, and grounding. It also evaluates only one manager-based framework, leaving alternative approaches, such as using decomposed questions as context for direct generation, unexplored.

\emph{Model Coverage.}
The experiments evaluate proprietary models from OpenAI (GPT-4o, GPT-4.1, and GPT-4.1 mini) and Anthropic (Claude Sonnet 4.6 and Claude Haiku 4.5). The findings may not generalize to open-weight models or Text-to-SQL-specialized systems, which may exhibit different capabilities and cost-performance characteristics.

\emph{Evaluation Limitations.}
Execution accuracy is the primary metric, supplemented by judge-gated evaluation for semantically equivalent outputs that fail strict comparison. However, execution against a single database instance cannot guarantee equivalence across all possible database states, and LLM-based judges may introduce additional evaluation uncertainty. Therefore, strict execution accuracy and judge-gated accuracy are reported separately.

\emph{Statistical Uncertainty.}
The benchmark contains 20 questions per SQL complexity category, with selective routing evaluated on smaller subsets. Despite paired evaluation on identical questions, the limited benchmark size reduces statistical power. The 95\% confidence intervals for paired differences span approximately $\pm$3--7 percentage points, limiting the ability to detect small effects. Thus, differences between configurations should be interpreted cautiously.

\emph{Computational Transparency and Model Accessibility.}
Because the evaluated models are accessed through commercial API's, their parameter counts and serving infrastructure are unavailable. We therefore report observable computational metrics, including API calls, latency, and model usage per pipeline stage.

%===============================
\section*{Ethical Considerations}

This work evaluates Text-to-SQL generation using publicly available datasets without personally identifiable information, including the FHWA's National Bridge Inventory and BIRD benchmarks. Human evaluation was limited to model-generated decomposition outputs, with no access to private databases or sensitive data. Experiments used commercial LLMs via API without transmitting confidential information. As generated SQL may contain errors, practical deployment should incorporate appropriate validation, particularly for high-impact applications. This work aims to advance reliability and responsible use of agentic Text-to-SQL systems.

\section*{Acknowledgments}
The authors acknowledge the Federal Highway Administration (FHWA) for providing public datasets that were essential to the testing and evaluation of this work.

% Custom bibliography entries only

\bibliographystyle{IEEEtran}
\bibliography{custom-IEEE}

\appendices

\section{Human Evaluation of Decomposition Quality}
\label{app:human}

Execution accuracy checks only the final SQL result, so it cannot show whether an error comes from decomposition or from later SQL generation. We therefore evaluated the generated sub-questions separately from SQL execution.

\begin{table}[!b]
\centering
\small
\begin{tabular}{lcccc}
\toprule
 & \multicolumn{2}{c}{\textbf{Question}} & \multicolumn{2}{c}{\textbf{Sub-question}} \\
\cmidrule(lr){2-3}\cmidrule(lr){4-5}
\textbf{Level} & $n/N$ & \% & $n/N$ & \% \\
\midrule
L1 Easy         & 5/5   & 100.0 & 12/12 & 100.0 \\
L2 Moderate     & 16/18 & 88.9  & 48/51 & 94.1  \\
L3 Intermediate & 19/20 & 95.0  & 68/69 & 98.6  \\
L4 Advanced     & 19/20 & 95.0  & 61/62 & 98.4  \\
L5 Expert       & 17/19 & 89.5  & 61/64 & 95.3  \\
\midrule
\textbf{All}    & \textbf{76/82} & \textbf{92.7} & \textbf{250/258} & \textbf{96.9} \\
\bottomrule
\end{tabular}
\caption{Human evaluation of decomposition quality by difficulty level. \emph{Question}: plans in which every sub-question was judged valid. \emph{Sub-question}: individually valid sub-questions. $n/N$ is count over total; \% is the corresponding rate. Two annotators reached 96.1\% raw agreement, with disagreements adjudicated by a third.}
\label{tab:decomp-quality}
\end{table}

The evaluation uses a forced decomposition run with a GPT-4.1-mini decomposer, which decomposed 82 of 100 questions into 258 sub-questions. These 82 questions cover all difficulty levels. Annotators independently labeled each sub-question as valid or invalid. A sub-question was valid if it was relevant and logically correct step toward answering the original question, regardless of whether its SQL executed. Annotators agreed on 248 of 258 labels, and a third annotator resolved the remaining disagreements.
Of the 258 sub-questions, 250 (96.9\%) are judged valid, and 76 of the 82 questions (92.7\%) contain valid sub-questions (Table~\ref{tab:decomp-quality}). The validity is greater than 94\% at each difficulty level, and the six questions with an invalid step were spread over L2 - L5. Invalid decomposition is unlikely to be primary error source. However, valid sub-questions may still be incomplete, misordered, or combined at the wrong scope during composition, leading to errors in final SQL composition.

\section{NBI Benchmark}
\label{app:Benchmark}

\emph{Database.}
The database is derived from the U.S. Federal Highway administration's National Bridge Inventory. The original release is a single wide table, which we split into six related tables with 1{,}197{,}381 rows and 34 columns in total (Table~\ref{tab:dbstats}). \smallcaps{BRIDGES} is the main table, \smallcaps{INSPECTIONS} records inspection events, and the other four are inspection tables linked by foriegn keys.

\emph{Questions.}
The benchmark questions were authored and verified by database experts. Each question was manually reviewed to verify alignment between the natural-language question, the intended analytical task, the gold SQL query, and the executed query results.

\emph{Coverage.}
The questions use all six tables and 32 of the 34 columns. \smallcaps{BRIDGES} appear in 96 queries, \smallcaps{STATES} appear in 43, \smallcaps{INSPECTIONS} in 23, \smallcaps{OWNER\_AGENCIES} in 9, \smallcaps{FUNCTIONAL\_CLASS} in 4 and \smallcaps{COUNTIES} in 5.

% NBI benchmark database statistics: per-table rows, columns, and how many
% of the 100 gold queries reference each table.

\begin{table}[!htbp]
\centering
\small
\begin{tabular}{lrrr}
\toprule
\textbf{Table} & \textbf{Rows} & \textbf{Cols} & \textbf{Queries} \\
\midrule
BRIDGES          & 569,864 & 17 & 96 \\
INSPECTIONS      & 624,193 &  7 & 23 \\
COUNTIES         &   3,216 &  4 &  5 \\
STATES           &      56 &  2 & 43 \\
OWNER\_AGENCIES  &      40 &  2 &  9 \\
FUNCTIONAL\_CLASS&      12 &  2 &  6 \\
\midrule
\textbf{Total}   & \textbf{1,197,381} & \textbf{34} & -- \\
\bottomrule
\end{tabular}
\caption{NBI benchmark database. \textbf{Queries} is the number of the
100 gold queries referencing each table; a query may reference several.}
\label{tab:dbstats}
\end{table}

\end{document}